\pdfoutput=1

\documentclass[11pt]{article}

\usepackage[dvipsnames]{xcolor}

\usepackage[preprint]{coling}

\usepackage{times}
\usepackage{latexsym}

\usepackage[T1]{fontenc}

\usepackage[utf8]{inputenc}

\usepackage{microtype}

\usepackage{inconsolata}

\usepackage{graphicx}

\usepackage{tikz}
\usetikzlibrary{math,calc,positioning,intersections,fit,matrix}
\usetikzlibrary{decorations.pathreplacing,decorations.markings,arrows.meta}
\usetikzlibrary{shapes.geometric}
\tikzset{>=latex}
\usetikzlibrary{shapes.misc}

\pgfdeclarelayer{background}
\pgfsetlayers{background,main}

\usepackage{booktabs}
\usepackage{amsfonts}
\usepackage{amsmath}
\usepackage{amssymb}
\usepackage{bm}
\usepackage[ruled]{algorithm2e}

\def\eqref#1{equation~\ref{#1}}

\def\1{\bm{1}}

\def\vzero{{\bm{0}}}
\def\vone{{\bm{1}}}

\def\va{{\bm{a}}}

\def\vc{{\bm{c}}}
\def\vd{{\bm{d}}}
\def\ve{{\bm{e}}}

\def\vp{{\bm{p}}}
\def\vq{{\bm{q}}}

\def\vs{{\bm{s}}}

\def\vw{{\bm{w}}}
\def\vx{{\bm{x}}}
\def\vy{{\bm{y}}}

\def\eva{{a}}

\def\evd{{d}}
\def\eve{{e}}

\def\evs{{s}}

\def\evw{{w}}

\def\mA{{\bm{A}}}
\def\mB{{\bm{B}}}
\def\mC{{\bm{C}}}
\def\mD{{\bm{D}}}

\def\mI{{\bm{I}}}

\def\mM{{\bm{M}}}

\def\mS{{\bm{S}}}

\def\mU{{\bm{U}}}

\def\mX{{\bm{X}}}

\def\mZ{{\bm{Z}}}

\DeclareMathAlphabet{\mathsfit}{\encodingdefault}{\sfdefault}{m}{sl}
\SetMathAlphabet{\mathsfit}{bold}{\encodingdefault}{\sfdefault}{bx}{n}

\def\emA{{A}}
\def\emB{{B}}

\def\emD{{D}}

\def\emM{{M}}

\def\emZ{{Z}}

\newcommand{\R}{\mathbb{R}}

\newcommand{\softmax}{\mathrm{softmax}}

\DeclareMathOperator*{\argmin}{arg\,min}

\newcommand{\tr}{\bm{\mathrm{tr}}}

\allowdisplaybreaks

\DeclareMathOperator{\rank}{rank}

\newtheorem{definition}{Definition}

\title{%
Few-Shot Domain Adaptation for Named-Entity Recognition\\
via Joint Constrained k-Means and Subspace Selection
}

\author{Ayoub Hammal\textsuperscript{{\normalfont 1}} \qquad Benno Uthayasooriyar\textsuperscript{{\normalfont 2,3}} \qquad Caio Corro\textsuperscript{{\normalfont 4}}
\\
\textsuperscript{1}Université Paris-Saclay, CNRS, LISN
\\
\textsuperscript{2}Data Analytics Solutions, SCOR
\quad
\textsuperscript{3}LMBA, CNRS, Université de Brest
\\
\textsuperscript{4}INSA Rennes, IRISA, Inria, CNRS, Université de Rennes
}

\begin{document}
\maketitle
\begin{abstract}
Named-entity recognition (NER) is a task that typically requires large annotated datasets, which limits its applicability across domains with varying entity definitions. This paper addresses few-shot NER, aiming to transfer knowledge to new domains with minimal supervision. Unlike previous approaches that rely solely on limited annotated data, we propose a weakly supervised algorithm that combines small labeled datasets with large amounts of unlabeled data. Our method extends the k-means algorithm with label supervision, cluster size constraints and domain-specific discriminative subspace selection. This unified framework achieves state-of-the-art results in few-shot NER on several English datasets.
\end{abstract}

\section{Introduction}

Named-entity recognition (NER) is a fundamental information retrieval task
that aims to identify entity mentions as well as their corresponding types in texts \cite{muc6,muc7ner}.
This problem can be tackled via standard structured prediction methods, \emph{e.g.}\ conditional random fields for segmentation \cite{lafferty2001crf,mccallum2003crfner,sarawagi2004semimarkov}.
As supervised learning approaches come at the expense of building large annotated datasets,
there is a growing interest in fine-tuning NER models using only a (very) small annotated dataset, called the support.

Specifically, we focus on the few-shot domain adaptation scenario:
a model is first trained on a large annotated dataset, and then fine-tuned on target data using as support only 1-5 examples per mention type.
We consider two different flavors of few-shot domain adaptation:
(1) tag set extension, \emph{i.e.}\ output domain transfer, where the model is fine-tuned to predict mention types that were previously unknown, but on the same input domain;
(2) input domain transfer, where the model is fine-tuned to predict mentions in previously unseen data sources, potentially using a different annotation scheme.
A natural approach in this setting is to build class prototypes from the support and rely on nearest neighbor classification for prediction \citep[][\emph{inter alia}]{fritzler-2019-fewshot-prototype,yang-2020-simple-fewshot,das-2022-container-contrastive}.


In this work, we propose a novel weakly-supervised learning method for few-shot NER that overcome limitations of previous work.
Firstly, our method is based on the k-means clustering algorithm which naturally benefits from access to extra unlabeled data that is often cheap and easy to collect.
Secondly, we introduce a ratio constraint on the number of words that are not part of mentions as extra learning information to make the most of unlabeled data, in the same spirit as \citet{effland-2021-expected-entity-ratio}.
To this end, we develop novel algorithms to take into account this ratio constraint in the training procedure.
Thirdly and lastly, we jointly learn a projection of the data into a subspace so that clusters are well separated, often referred to as discriminative clustering \cite{ding-2007-adaptative-dimension-reduction,ye-2007-discriminative-k-means}.
All in all, our procedure is grounded on a well-defined training problem and efficient optimization algorithms.

A well-known issue of few-shot learning is the instability of training, \emph{i.e.}\ there can be a high variance between runs of the same training process for the same support, mainly due to source of randomness.
To fix this issue, we devise a strictly deterministic training procedure, meaning that two runs will lead to the same results, 
as there not a single call to a random number generator.
We achieve this by using batch updates (\emph{i.e.}\ on the full training objective) instead of a stochastic optimization algorithm that operates on minibatches \cite{bottou2010largescale}.
Moreover, we propose a deterministic initialization procedure for our k-means in order to bypass the usual random initialization.
Finally, note that our training algorithm is based on a parameter-free optimization method, meaning that there is no learning parameter to tune. 

Our contributions can be summarized as follows:
\begin{itemize}
    \item We formalize few-shot domain adaptation as a k-means clustering problem, which can benefit from extra unlabeled data;
    \item We propose novel algorithms for the E-step of k-means that allows to introduce ratio constraints in both hard and soft variants;
    \item We extend the clustering process to jointly project the data into a subspace where clusters are well-separated;
    \item We evaluate our approach in different few-shot settings and achieve novel state-of-the-art results compared to previous work.
\end{itemize}
Our code is publicly available.\footnote{\url{https://github.com/ayoubhammal/ckss4ner}}

\textbf{Notation.}
We write scalars (resp.\ sets) in lowercase (resp.\ uppercase),
and vectors (resp. matrices) in bold lowercase (resp.\ uppercase).
Given a matrix $\mM$, we denote $\emM_{ij}$ the element at row $i$ and column $j$, and $\mM_i$ the vector corresponding to row $i$.
We denote $\| \va \| = \sqrt{\sum_i \eva_i^2}$ the L2 norm, $\langle \mA, \mB\rangle = \sum_{i, j} \emA_{ij}\emB_{ij}$ the sum of entries of the Hadamard product (\emph{i.e}.\ dot product if arguments are vectors)
and $\tr(\mM) = \sum_i \emM_{ii}$ the trace.
Given $i \in \mathbb N_{++}$, we write $[i]$ the set $\{1..i\}$.
Given a set $S$, we write $\mathcal P(S)$ the powerset of $S$ and $\mathcal P_i(S)$ the set of all subsets of $S$ with cardinality $i$.
We denote $\triangle(k) = \{\va \in \R^k_+ | \sum_i \eva_i = 1\}$ the simplex of dimension $k-1$.
\section{Few-Shot Named-Entity Recognition}

The NER problem aims to identify entity mentions in texts.
This chunking task is often reduced to a word tagging problem using the \textsc{BIO} scheme \citep{ramshaw1995bio}:
each word is tagged either with \textsc{O} (not in a mention), \textsc{B-Type} (first word of a mention) or \textsc{I-Type} (following words in a mention), where $\textsc{Type}$ is any allowed mention type (\emph{e.g.}\ \textsc{Loc}, \textsc{Org}, etc.)
Following previous work in the few-shot scenario \cite{yang-2020-simple-fewshot,das-2022-container-contrastive}, we rely on a simplified \textsc{IO} scheme, where each word is either tagged with \textsc{O} or \textsc{I-Type}, for example:\newline
\begin{tikzpicture}[
    every node/.style={
        rectangle,
        inner xsep=0cm,
        inner ysep=0.1cm,
        text height=1.5ex,
        text depth=.25ex,
    }
]
    \node (un) [rectangle] {\texttt{U.N.}};
    \node (official) [rectangle, right=0.2cm of un] {\texttt{official}};
    \node (ekeus) [rectangle, right=0.2cm of official] {\texttt{Ekeus}};
    \node (heads) [rectangle, right=0.2cm of ekeus] {\texttt{heads}};
    \node (for) [rectangle, right=0.2cm of heads] {\texttt{for}};
    \node (baghdad) [rectangle, right=0.2cm of for] {\texttt{Baghdad}};

    \node (t_un) [rectangle, below=-0.1cm of un] {\textsc{I-Org}};
    \node (t_official) [rectangle, below=-0.1cm of official] {\textsc{O}};
    \node (t_ekeus) [rectangle, below=-0.1cm of ekeus] {\textsc{I-Per}};
    \node (t_heads) [rectangle, below=-0.1cm of heads] {\textsc{O}};
    \node (t_for) [rectangle, below=-0.1cm of for] {\textsc{O}};
    \node (t_baghdad) [rectangle, below=-0.1cm of baghdad] {\textsc{I-Loc}};
\end{tikzpicture}    
We write $T = \{\textsc{O}, \textsc{I-Loc}, ...\}$ the set of tags.

\textbf{Weighting model.}
Let $\vs = (\evs_1, \dots, \evs_n)$ be an input sentence of $n$ words, which is passed through a neural network that computes $d$-dimensional hidden representations $(\vx^{(1)}, \dots, \vx^{(n)})$, \emph{e.g.}\ \textsc{Bert} \citep{devlin2019bert}.
Then, output tag weights for word $i$ can be computed via a linear model:
\begin{align}
    \vw^{(i)} = \mB \vx^{(i)} + \vd
    \label{eq:weights_linear}
\end{align}
where $\vw^{(i)} \in \R^{|T|}$ are output weights, $\mB \in \R^{|T| \times d}$ and $\vd \in \R^{|T|}$ are the model parameters.
The prediction is simply the tag of maximum weight.
In our setting, we instead use tag prototypes so that we can rely on clustering for learning.
As such, the weight of a tag is proportional to the negative squared Euclidean distance with the prototype:
\begin{align}
\evw^{(i)}_j = - \frac{1}{2}\| \vx^{(i)} - \vc^{(j)} \|^2 \label{eq:weights_protptype}
\end{align}
where $\vc^{(j)} \in \R^d$ is the \emph{prototype} of tag $j \in [~|T|~]$.%
\footnote{Models (\ref{eq:weights_linear}) and (\ref{eq:weights_protptype}) are equivalent, see Appendix~\ref{app:linear_prototype}.}

In practice, it can be useful to have several prototypes per tag,
in particular for the \textsc{O} tag that gathers heterogeneous classes of words.
Let $k$ be the total number of prototypes and $\mC \in \R^{k \times d}$ be the matrix that contains in each row a prototype of dimension $d$.
Let $\phi: [k] \to T$ be the function that assigns to each prototype a tag and $\phi^{-1}$ its preimage function:
\[
\phi^{-1}(t) = \{ i \in [k]~|~\phi(i) = t \}\,.
\]
The weight of tag $t \in T$ for word $\evs^{(i)}$ is defined as:
\begin{align*}
    - \min_{j \in \phi^{-1}(t)}~\frac{1}{2}\| \vx^{(i)} - \mC_j \|^2\,,
\end{align*}
or, in other words, the weight of a given tag $t \in T$ depends on the closest prototype according to $\phi^{-1}$.

\textbf{Few-shot evaluation.}
We simulate a transfer learning scenario as described by \citet{yang-2020-simple-fewshot}:
we first pre-train a model on a \emph{source domain} for which there exists a large annotated dataset,
and then fine-tune the model on a \emph{target domain} using only a few labeled sentences, called the \emph{support}.
The target domain may have different mention labels than the source domain.\footnote{This means that the output layer used during pre-training, \emph{i.e.}\ either Equation~(\ref{eq:weights_linear}) or~(\ref{eq:weights_protptype}), is not used for the target domain.}
However, contrary to \citet{yang-2020-simple-fewshot}, we assume access to a large unlabeled dataset in the target domain, which is often easy and cheap to obtain.

For pre-training, we simply use the negative log-likelihood loss defined as:
\[
\ell(\vw ; \vy) =
- \langle \vw, \vy \rangle + \log \sum_i \exp \evw_i\,,
\]
where $\vy$ is a one-hot vector indicating the gold tag.
Contrary to previous works \citep{fritzler-2019-fewshot-prototype, yang-2020-simple-fewshot,das-2022-container-contrastive}, we compute tag weights using Equation~(\ref{eq:weights_protptype}) 
instead of~(\ref{eq:weights_linear}), as it is more similar to the fine-tuning method.
\footnote{It led to better results in early experiments.}

\textbf{Fine-tuning data.}
We will denote the fine-tuning dataset as the matrix $\mX \in \R^{n \times d}$, where $n$ is total number of words in the fine-tuning data (labeled and unlabeled). 
Moreover, the matrix $\mZ \in \{0, 1\}^{n \times k}$ indicates which prototypes are allowed for each word.
In other words, $\emZ_{ij}$ is equal to 1 if and only if prototype $j$ is allowed for datapoint $i$.
That is, if $i \in [n]$ is tagged with $t \in T$ (\emph{i.e.}\ it is in the support), then:
\[
\emZ_{ij} = \begin{cases}
    1 \quad&\text{if } j \in \phi^{-1}(t)\,,\\
    0 \quad&\text{otherwise.}
\end{cases}
\]
Otherwise, if word $i' \in [n]$ is not tagged (\emph{i.e.}\ it is not in the support), then $\emZ_{i'j} = 1, \forall j \in [k]$.
\section{Weakly-Supervised Clustering}

In our few-shot settings, we have access to an additional unlabeled dataset, that is, we learn from low-recall data where only a few mentions are annotated.
In this setting, it is useful to introduce extra knowledge during training as a supervision signal.
We follow  \citet{effland-2021-expected-entity-ratio} and impose a ratio constraint on the \textsc{O} tag during training.

We first recall the k-means algorithm using our notation.
Then, we propose novel algorithms for the E step that allows to enforce the ratio constraint.
Finally, we explain an initialization strategy that leads to fully deterministic fine-tuning method.

\subsection{The k-Means Algorithm}

Clustering aims to find a partition of $\mX$ into $k$ clusters that minimizes the intra-cluster dispersion:
\begin{align}
    \min_{\pi \in \mathcal P_k(\,[n]\,)}
    \quad
    \sum_{C \in \pi}
    \sum_{i \in C}
    \| \mX_i - \overline{C} \|^2\,,
\end{align}
where $\overline{C} = |C|^{-1}\sum_{i \in C} \mX_i$ denotes the cluster centroid.
It can be shown that this is a NP-hard combinatorial problem \citep{dasgupta2008hardness,aloise2009hardness}.
The main ideas behind the k-means algorithm are:
(1)~allow clusters to contain no datapoint;
(2)~transform the combinatorial search $\pi \in \mathcal P_k(\,[n]\,)$ into a continuous problem over a cluster assignation matrix $\mA \in \R^{n \times k}$;
(3)~replace the dispersion around a cluster centroid by its variational formulation:
\begin{align}
\sum_{i \in C} \| \mX_i - \overline{C} \|^2
=
\min_{\vc \in \R^d} \sum_{i \in C} \| \mX_i - \vc \|^2\,.
\label{eq:var_dist}
\end{align}
We obtain the following optimization problem:
\begin{align}
    \min_{\mA, \mC}~ 
    & \sum_{i \in [n]} \sum_{j \in [k]} \emA_{ij} \| \mX_i - \mC_j\|^2
    + \Omega(\mA)\,,
    \nonumber
    \\
    \text{s.t.}~
    & \sum_{j=1}^k \emA_{ij} = 1 \quad \forall i \in [n]\,,  \label{cst:simplex}
    \\
    & \mA \in \R^{n \times k}_+
    \text{~and~} \mC \in \R^{k \times d}\,, \label{cst:vars}
    \\
\intertext{%
where $\Omega(\cdot)$ is a regularizer that can be interpreted in a similar way to the regularizer in the Fenchel-Young losses framework \citep{blondel2020fy}.
Equation (\ref{cst:simplex}) ensures that each datapoint is assigned to exactly one cluster.
In our setting, cluster centroids in $\mC$ corresponds to (learned) prototypes.
To take into account the supervision knowledge encoded in $\mZ$, we add the following constraint:
}
    & \emA_{ij} \leq \emZ_{ij} \quad \forall i \in [n]\,, j \in [k]\,, \label{cst:supervision}
\end{align}
\emph{i.e.}\ if a cluster is forbidden for a datapoint, the corresponding value in $\mA$ is forced to 0.

The objective of the k-means problem is non-convex \citep[][Eq.~3.1]{an2006dc_clustering}, however it is bi-convex, that is convex in $\mA$ (resp.\ $\mC$) when $\mC$ is fixed (resp.\ $\mA$).
Therefore, the standard optimization method is based on Alternate Convex Search \citep[][Sec.~5.9]{hastie2015sparsity}:
we iteratively minimize the objective over $\mA$ (E step) and over $\mC$ (M step).\footnote{E and M names comes from the EM algorithm, of which k-means is a special case \citep[][Sec.~14.3.7]{hastie2009ml}.}
Importantly, this optimization procedure is not specific to the original k-means and applies to all problems with the same properties, \emph{e.g.}\ when adding Constraint~(\ref{cst:supervision}) and the ratio Constraint~(\ref{cst:ratio}).

\textbf{Hard k-means.}
If $\Omega(\mA) = 0$, we obtain the standard hard k-means problem.
\textbf{(E step)} Note that if we minimize over $\mA$ only, the cluster distance term is constant. Let $\mD$ be a matrix s.t.\ $\emD_{ij} = \| \mX_i - \mC_j \|^2$, then the optimal assignation is:
    \begin{align}
    \label{eq:e_step}
    \hspace{-0.2cm}\widehat{\mA}_{i}
    \in
    \argmin_{\ve \in \triangle(k)}
    ~\langle \ve, \mD_i \rangle
    ~\text{s.t.}~
    \eve_j \leq \emZ_{ij}, \forall j \in [k]\,,
    \end{align}
    where it is usual to choose one of optimal simplex corners in case of ties.
    This problem can be solved in $\mathcal O(k)$ for single datapoint, hence the time-complexity is $\mathcal O(nk)$.
\textbf{(M step)} Minimizing over cluster centroids $\mC$ simply yields:
    \begin{align}
        \label{eq:m_step}
        \widehat{\mC}_j
        = \frac{\sum_{i} \emA_{ij} \mX_i}{\sum_{i} \emA_{ij}}\,.
    \end{align}
    Time-complexity is $\mathcal O(nk)$.

\textbf{Soft k-means.}
If $\Omega(\mA)$ is the negative Shannon entropy defined as follows:
\[
    \Omega(\mA) = - H(\mA) = \langle \mA, \log \mA \rangle - \langle \mA, \1 \rangle\,,
\]
the resulting algorithm is known as soft k-means.\footnote{Not to be confused with fuzzy clustering that optimizes a different objective \citep{dunn-1973-fuzzy, bezdek-1981-pattern-recognition-fuzzy-objective}.}
The optimal solution of the E step becomes:
\begin{align*}
    \widehat{\emA}_{ij} = \frac{
            \emZ_{ij} \exp \emD_{ij}
        }{
            \sum_{j'} \emZ_{ij'} \exp \emD_{ij'}
        }\,,
\end{align*}
see \citep[Ex.~3.71]{beck2017opt}.
It can be computed in $\mathcal O(nk)$.
The M step is left unchanged.

\subsection{Weak Supervision via Ratio Constraint}
\begin{figure*}
\centering
\begin{minipage}[b]{0.56\textwidth}
    \centering
    \vspace{0cm}
    \begin{tikzpicture}[
    vertex/.style={
        circle,
        fill=black,
        inner sep=0.15em
    },
    weight/.style={
        midway,
        pos=0.22,
        fill=white,
        font=\small,
        inner sep=0.1em
    },
    scale=0.8
]

\node[vertex,label={left:\small $\vx^{(1)}$}] at (0, 2.5) (x1) {};
\node[vertex,label={left:\small $\vx^{(2)}$}] at (0, 0) (x2) {};
\node[vertex,label={left:\small $\vx^{(3)}$}] at (0, -2.5) (x3) {};

\node[vertex,label={right:\small\textsc{I-Org}}] at (3, 2.25) (i1) {};
\node[vertex,label={right:\small\textsc{I-Loc}}] at (3, 0.75) (i2) {};
\node[vertex,label={right:\small\textsc{O}}] at (3, -0.75) (o1) {};
\node[vertex,label={right:\small\textsc{O}}] at (3, -2.25) (o2) {};

\draw[black,dashed,rounded corners] ($(i1.north west)+(-0.3,0.3)$)  rectangle ($(i2.south east)+(1.5,-0.3)$);
\path ($(i1.north east)+(+1.5,-.3)$) --  ($(i2.south east)+(1.5,-0.3)$) node[midway,above,rotate=-90,xshift=-0.6em] {Other clusters};

\draw[black,dashed,rounded corners] ($(o1.north west)+(-0.3,0.3)$)  rectangle ($(o2.south east)+(1.5,-0.3)$);
\path ($(o1.north east)+(1.5,0.3)$) --  ($(o2.south east)+(1.5,-0.3)$) node[midway,above,rotate=-90] {\textsc{O} clusters};

\draw[-] (x1) -- node[weight] {$4$} (i1);
\draw[-] (x1) -- node[weight] {$1$} (i2);
\draw[-] (x1) -- node[weight] {$5$} (o1);
\draw[-] (x1) -- node[weight] {$6$} (o2);

\draw[-] (x2) -- node[weight] {$3$} (i1);
\draw[-] (x2) -- node[weight] {$9$} (i2);
\draw[-] (x2) -- node[weight] {$2$} (o1);
\draw[-] (x2) -- node[weight] {$4$} (o2);

\draw[-] (x3) -- node[weight] {$8$} (i1);
\draw[-] (x3) -- node[weight] {$4$} (i2);
\draw[-] (x3) -- node[weight] {$2$} (o1);
\draw[-] (x3) -- node[weight] {$7$} (o2);

\node[vertex,label={left:\small $\vx^{(1)}$}] at (7, 1.5) (cx1) {};
\node[vertex,label={left:\small $\vx^{(2)}$}] at (7, 0) (cx2) {};
\node[vertex,label={left:\small $\vx^{(3)}$}] at (7, -1.5) (cx3) {};

\node[vertex,label={right:\small\textsc{Others}}] at (10, 1) (cothers) {};
\node[vertex,label={right:\small\textsc{O}}] at (10, -1) (co) {};

\draw[-, ultra thick, red] (cx1) -- node[weight] {$1$} (cothers);
\draw[-] (cx1) -- node[weight] {$5$} (co);

\draw[-, ultra thick, red] (cx2) -- node[weight] {$3$} (cothers);
\draw[-] (cx2) -- node[weight] {$2$} (co);

\draw[-] (cx3) -- node[weight] {$4$} (cothers);
\draw[-, ultra thick, red] (cx3) -- node[weight] {$2$} (co);


\coordinate (ctop_n1) at ($(i1.north west)+(-0.3,0.3)$);
\coordinate (ctop_n2) at ($(i1.north east)+(1.5,0.3)$);
\coordinate (ctop_n1bis) at ($(ctop_n1)!0.5!(ctop_n2)+(0,0.6)$);
\coordinate (ctop_n2bis) at ($(cothers.north)+(0,0.1)$);

\draw[-{Triangle[width=8pt,length=6pt]}, line width=2pt]
($(ctop_n1bis)-(0,0.5)$)
--
(ctop_n1bis)
-|
(ctop_n2bis)
;

\draw (ctop_n1bis) -- (ctop_n1bis -|ctop_n2bis) node[midway, above] {\small contract};


\coordinate (cbot_n1) at ($(o2.south west)+(-0.3,-0.3)$);
\coordinate (cbot_n2) at ($(o2.south east)+(1.5,-0.3)$);
\coordinate (cbot_n1bis) at ($(cbot_n1)!0.5!(cbot_n2)+(0,-0.6)$);
\coordinate (cbot_n2bis) at ($(co.south)+(0,-0.1)$);

\draw[-{Triangle[width=8pt,length=6pt]}, line width=2pt]
($(cbot_n1bis)+(0,0.5)$)
--
(cbot_n1bis)
-|
(cbot_n2bis)
;

\draw (cbot_n1bis) -- (cbot_n1bis -|cbot_n2bis) node[midway, below] {\small contract};

\end{tikzpicture}
\caption{%
Illustration of the E step with ratio constraints.
In the two bipartite graphs, left (resp.\ right) nodes represents datapoints (resp.\ clusters).
The left graph is the full graph, where edge weights indicate distances between nodes and clusters.
By contracting the two sets of clusters, we obtain a new graph,
on which we can run the E step with a ratio constraint for the contracted \textsc{O} cluster (ratio is set to $1/3$ in the example).
Thick red edges indicate the optimal solution.
Note that, without the ratio constraint, $\vx^{(2)}$ would be assigned to the \textsc{O} cluster.
}
\label{fig:e_step_ratio}
\end{minipage}\hfill%
\begin{minipage}[b]{0.41\textwidth}
    \centering
    \vspace{0cm}
    \begin{tikzpicture}[
    c1/.style={
        circle,
        inner sep=0.15em,
        fill=red
    },
    c2/.style={
        cross out,
        draw, 
        minimum size=2*(#1-\pgflinewidth), 
        inner sep=0pt,
        outer sep=0pt,
        line width=1pt,
        rotate=0
    },
    c2/.default={3pt},
    c3/.style={
        star,
        star points=5,
        star point height=0.2em,
        fill=ForestGreen,
        inner sep=0.08em,
        draw=black,
        line width=0.02em
    },
]

\node[c1] (n1) at (0.5  ,  0.375) {};
\node[c1] (n2) at (1.   ,  0.875) {};
\node[c2] (n3) at (0.   ,  0.375) {};
\node[c2] (n4) at (-1.5  , -1.625) {};

\draw[black,dashed,rounded corners] ($(n4.south west)+(-0.3,-0.3)$)  rectangle ($(n2.north east)+(0.3,0.3)$);

\node[c3] (n_unk) at (-0.1  ,  0.275) {};

\node[c1] (pn1) at (-1.76690442, -4) {};
\node[c1] (pn2) at (-2.76078815 ,  -4) {};
\node[c2] (pn3) at (2.31906205 ,  -4) {};
\node[c2] (pn4) at (2.20863052 , -4) {};

\node[c3] (p_n_unk) at (2.51783879 , -4) {};

\begin{pgfonlayer}{background}
\draw[black,dashed] ($(pn2)+(-0.3,0)$)  -- ($(p_n_unk)+(0.3,0)$);
\end{pgfonlayer}

\coordinate (arrow_anchor) at ($(pn2)!0.5!(p_n_unk)$);

\draw[-{Triangle[width=8pt,length=6pt]}, line width=2pt]
($(arrow_anchor)+(0,1.9)$)
--
node[midway,right] {\small projection}
($(arrow_anchor)+(0,0.1)$)
;

\end{tikzpicture}

\caption{%
Illustration of the benefit of subspace selection.
\textbf{(top)} Data in its original 2D space.
We assume the constrained clustering results in two clusters:
one containing the two black crosses and the other containing the two red circles.
Let the green star be a test point.
Intuitively, it should be classified in the black crosses cluster, however, it is closer to the other cluster centroid!
\textbf{(bottom)}
Data after projection in a 1D space.
The test point is now correctly classified.
}
\label{fig:subspace}
\end{minipage}\hfill
\end{figure*}

Let $r_\textsc{O}$ be the expected ratio of words tagged with $\textsc{O}$.
To use this extra information, we can add the following constraint to the clustering problem:
\begin{align}
    \sum_{i \in [n]}\sum_{j \in \phi^{-1}(\textsc{O})} \emA_{ij} = n \times r_{\textsc{O}}\,.
    \label{cst:ratio}
\end{align}
Note that this constraint only applies to the assignment matrix $\mA$, therefore it only impacts the E step.
In the following, we propose novel algorithms to compute the E step with Constraint~(\ref{cst:ratio}).

\textbf{Hard k-means.}
The (unconstrained) E step can be seen as a graph problem:
(1) construct a bipartite graph where one set of nodes corresponds to datapoints and the other to clusters, and there is an edge connecting a datapoint $i$ with a cluster $j$ with weight $\emD_{ij}$ if and only if $\emZ_{ij} = 1$, see Figure~\ref{fig:e_step_ratio} (left);
(2) compute the one-to-many assignment of minimum weight, where each node representing a datapoint is assigned to exactly one cluster (\emph{i.e.}\ it has exactly one incident edge in the solution).

To compute the solution in the constrained case, note that we can focus solely on whether a datapoint is assigned to one of the \textsc{O} clusters or not.
That is, we can divide cluster nodes into two groups: nodes representing $\textsc{O}$ clusters (\emph{i.e.}\ elements of $\phi^{-1}(\textsc{O})$) and nodes representing other clusters.
We can therefore \emph{contract} each group into a single node, where we keep only the edge of minimum weight between a datapoint and nodes in the contracted group, and compute the solution of the constrained E step on this simpler graph, see Figure~\ref{fig:e_step_ratio} (right).
It is trivial to construct the solution for the original graph from a contracted graph solution.

To this end, we build vectors $\vd^{(\textsc{O})} \in \R^n$ (resp.\ $\vd^{(\textsc{Others})} \in \R^n$) that indicates the distance between each datapoint and its closest \textsc{O} cluster (resp.\ non-\textsc{O} cluster), that is:
\begin{align*}
&\evd^{(\textsc{O})}_i &&= \hspace{-0.5em} \min_{j \in \phi^{-1}(\textsc{O})} \hspace{-0.5em} \emD_{ij}
\quad&&\text{s.t.}~~\emZ_{ij} = 1
\\
\text{and}\quad
&\evd^{(\textsc{Others})}_i &&= \hspace{-0.5em} \min_{j \notin \phi^{-1}(\textsc{O})} \hspace{-0.5em} \emD_{ij}
\quad&&\text{s.t.}~~\emZ_{ij} = 1\,,
\end{align*}
where the minimum is set to $\infty$ if the search space is empty.

Let $\va \in \{0, 1\}^n$ be an assignation vector to the $\textsc{O}$ group of clusters, \emph{i.e.}\ $\eva_{i} = 1$
if and only if $\vx^{(i)}$ is assigned to a \textsc{O} cluster.
Then, there is a one-to-one mapping between solutions $\widehat{\mA}$ of the constrained E step and solutions $\widehat{\va}$ of the following problem:
\begin{align*}
    \min_{\va}~& \langle \va, \vd^{(\textsc{O})} \rangle + \langle 1-\va, \vd^{(\textsc{Others})} \rangle
    \\
    \text{s.t.}~
    & \sum_{i \in [n]} \eva_i = n \times r_\textsc{O}
    \quad\text{and}\quad\va \in \{0, 1\}^n\,.
\end{align*}
The objective can be rewritten as follows:
\begin{align*}
    &\langle \va, \vd^{(\textsc{O})} \rangle + \langle 1-\va, \vd^{(\textsc{Others})} \rangle
    \\
    &=
    \langle \va, \underbrace{\vd^{(\textsc{O})} - \vd^{(\textsc{Others})}}_{= \vd'} \rangle + \underbrace{\langle \vone, \vd^{(\textsc{Others})} \rangle}_{\text{constant}}\,,
\end{align*}
where the second term is constant.
Computing the optimal $\widehat{\va}$ is therefore reduced to find the $n \times r_\textsc{O}$ smallest values in the penalized distance vector $\vd'$.
It is then trivial to build $\widehat{\mA}$ from $\widehat{\va}$ by inspecting which edges was kept in the contraction step.
Time-complexity is $\mathcal O(nk + n \log n)$ since it requires a partial sort of $\vd'$.


\textbf{Soft k-means.}
Constraints (\ref{cst:simplex}), (\ref{cst:supervision}) and (\ref{cst:ratio}) can be rewritten as inclusion in the intersection of the following affine subspaces:
\begin{align*}
    S^{(1)} &= \left\{ \mA \in \R^{n \times k}_+ ~\middle|~ \forall i, j: \emZ_{ij} = 0 \Leftrightarrow \emA_{ij} = 0 \right\}
    \\
    S^{(2)} &= \left\{ \mA \in \R^{n \times k}_+ ~\middle|~ \mA \vone = \vone \right\}
    \\
    S^{(3)} &= \left\{ \mA \in \R^{n \times k}_+ ~\middle|~ \sum_{i\in [n]} \sum_{j \in \phi^{-1}(\textsc{O})} \emA_{ij} = n \times r_\textsc{O} \right\}
\end{align*}
\emph{i.e.}\ $\mA \in S^{(1)} \cap S^{(2)} \cap S^{(3)}$.
We can rewrite the E step as a KL projection into this intersection:
\begin{align*}
    &\argmin_{\mA}~\langle \mA, \mD \rangle - H(\mA)~\text{s.t.}~(\ref{cst:supervision}), (\ref{cst:simplex}) \text{~and}~(\ref{cst:ratio})
    \\
    = & \argmin_\mA~KL[\mA | \exp(-\mD)]~\text{s.t.}~\mA \in \bigcap_{i=1}^3 S^{(i)}
\end{align*}
This problem can be (approximatively) solved using the iterative Bregman projection algorithm \citep{bregman1967proj,censor1997parallel},
which have recently been popular in the optimal transport literature \cite{benamou2015bregmanot}.
We iteratively project the current estimate into $S^{(1)} \cap S^{(2)}$ and $S^{(1)} \cap S^{(3)}$.
More details are given in Appendix~\ref{app:appendix-constrained-soft}.

\subsection{Initialization}

It is well known that k-means solution heavily depends on initialization \cite{bradley1998refining}, and several runs with different random initializations may produce quite different results.
Therefore, we opt for a deterministic approach for cluster center initialization.
An important advantage of our approach is that it improves reproducibility.

We assume that for each tag $t \in T$, there exists at least $|\phi^{-1}(t)|$ words annotated with $t$ in the dataset.
If $|\phi^{-1}(t)| = 1$,
we initialize the cluster centroid as the average of hidden representation of words labeled with $t$.
Otherwise,
we rely on greedy agglomerative hierarchical clustering using the Ward linkage strategy \cite[][Sec.\ 10.9]{duda2000patternclassification}.
We cut the dendrogram to obtain $|\phi^{-1} (t)|$ clusters whose centroids will serve as initial centroids for the k-means procedure.

\section{Subspace Selection}

Although constrained clustering is convenient for weakly-supervised few-shot learning,
it can lead to problems inherent to the clustering assumption:
the property that each datapoint is assigned to its closest neighbor may not be satisfied in the training data due to ratio or supervision constraints.
At test time, this may result in incorrect predictions.
To bypass this issue, we jointly learn a transformation of the data so that clusters are well separated, see Figure~\ref{fig:subspace}.
We focus on subspace selection, \emph{i.e.}\ the transformation is restricted to a linear projection.

\subsection{Problem Definition}

Let $\mU^\top \in \R^{p \times d}$, be a projection matrix of rank $p \leq d$.
Then, $\vx' = \mU^\top \vx \in \R^p$ is a projection of $\vx \in \R^d$ into the p-dimensional subspace defined by the linear map.\footnote{Note that this is equivalent to defining a custom metric
$\| \mU^\top \vx - \mU^\top\vc \|^2
= (\vx - \vc)^\top \mU \mU^\top(\vx - \vc)
    = \| \vx - \vc \|^2_{\mU \mU^\top}$,
called the Mahalanobis distance parameterized by $\mU \mU^\top$.}
The new joint constrained clustering an subspace selection problem is:
\begin{align*}
    \min_{\mA, \mC, \mU} \; 
    & \sum_{i \in [n]} \sum_{j \in [k]} \emA_{ij} \| \mU^\top \mX_i - \mU^\top \mC_j\|^2
    + \Omega(\mA)\,
    \\
    \text{s.t.} \quad &(\ref{cst:simplex}), (\ref{cst:vars})\text{~and~}(\ref{cst:supervision}).
\end{align*}
One issue with this problem formulation is that it has a trivial but non interesting global optimum for the first term by setting $\mU = \vzero$, \emph{i.e.}\ collapsing all points, as there is no constraint on $\mU$.

Given a cluster assignment matrix $\mA$, we assume $\widehat{\mC}$ are the optimal centroids given by Equation~(\ref{eq:m_step}).
To simplify notation, the dependency on $\mA$ of $\widehat{\mC}$ is not explicitly written.
The total, within-class and between-class scatter (correlation) matrices are:
\begin{align*}
    \mS_{\mU}^{\text{(t)}}
    &= \sum_{i \in [n]} (\mU^\top\mX_i - \mU^\top\overline{\vx}) (\mU^\top\mX_i - \mU^\top\overline{\vx})^\top
    \\
    \mS_{\mU, \mA}^{\text{(w)}}
    &= \sum_{j \in [k]}\sum_{i\in n[n]} \emA_{ij} \begin{array}[t]{l}(\mU^\top\mX_i - \mU^\top\widehat{\mC}_j) \\ \times (\mU^\top\mX_i - \mU^\top\widehat{\mC}_j)^\top \end{array}
    \\
    \mS_{\mU, \mA}^{\text{(b)}}
    &= \sum_{j \in [k]} \sum_{i\in n[n]} \emA_{ij} \begin{array}[t]{l}(\mU^\top\widehat{\mC}_j - \mU^\top\overline{\vx}) \\ \times (\mU^\top\widehat{\mC}_j - \mU^\top\overline{\vx})^\top \end{array}
\end{align*}
where $\overline{\vx} = n^{-1}\sum_{i \in [n]} \mX_i$ is the sample mean.
$\tr(\mS_{\mU}^{\text{(t)}})$, $\tr(\mS_{\mU, \mA}^{\text{(w)}})$ and $\tr(\mS_{\mU, \mA}^{\text{(b)}})$ correspond to data, intra-cluster and inter-cluster dispersion.

The following equalities holds (Appendix~\ref{app:scatter_eq}):
\begin{align}
    \label{eq:scatter_eq}
    \mS_{\mU}^{\text{(t)}} &= \mS_{\mU, \mA}^{\text{(w)}} + \mS_{\mU, \mA}^{\text{(b)}}
    \\
    \label{eq:dispersion_eq}
    \tr(\mS_{\mU}^{\text{(t)}}) &= \tr(\mS_{\mU, \mA}^{\text{(w)}}) + \tr(\mS_{\mU, \mA}^{\text{(b)}})
\end{align}
If $\mU$ \emph{is fixed to the identity matrix} $\mI$ (\emph{i.e.}\ no learned subspace selection), the left-hand side is constant as it does not depend on the clustering: minimizing the intra-cluster dispersion is equivalent to maximizing the inter-cluster dispersion, so there is no cluster collapse.
When jointly learning $\mU$, we propose to fix the expected data dispersion as follows:\footnote{See Appendix~\ref{app:scatter_proj} for proof of the equivalence.}
\begin{align}
    \mS^{\text{(t)}}_{\mU} = \mI
    \quad\Leftrightarrow\quad\mU^\top \mS^{\text{(t)}}_{\mI} \mU = \mI. \label{eq:cst_dispersion}
\end{align}
which prevent data and clusters collapse.

\subsection{Optimization Algorithm}

We follow an alternate convex search procedure where variables are visited in order $\mA \to \mC \to \mU$.
Minimizing over $\mA$ requires to take into account for the projection when computing matrix~$\mD$ in Equation~(\ref{eq:e_step}), \emph{i.e.}\ we set $\emD_{ij} = \| \mU^\top \mX_i - \mU^\top\mC_j\|^2$.
Optimization over $\mC$ is left unchanged, see App.~\ref{app:m_step_subspace}.

We are left with optimization over $\mU$.
Similarly to Equation~(\ref{eq:cst_dispersion}), we can rewrite the objective as:
\[
    \tr(\mU^\top \mS_{\mI, \mA} \mU) + \Omega(\mA)\,.
\]
Ignoring the constant term, the Lagrangian is:
\begin{align*}
    \mathcal L(\mU, \bf\Lambda)
    = &\tr (\mU^\top \mS_{\mI, \mA}^{(w)} \mU) \\[-0.17cm]
    &- \tr ({\bf\Lambda}^\top (\mU^\top \mS_{\mI}^{(t)} \mU - \mI)),
\end{align*}
where ${\bf\Lambda} \in \R^{p \times p}$ are dual variables associated with Constraint~(\ref{eq:cst_dispersion}), a.k.a.\ Lagrangian multipliers.
${\bf\Lambda}$ is implicitly constrained to be diagonal at optimality \citep[][App.~B]{ghojogh-2023-eigenvalue-generalized-eigenvalue-problems}.
By stationarity (\emph{i.e.}\ differentiating $\mathcal L$ w.r.t.\ $\mU$), a primal-dual pair of variable $\widehat \mU$ and $\widehat{\bf\Lambda}$ are minimizer if and only if:
\begin{align}
    \label{eq:stationarity}
    \mS_{\mI, \mA}^{(w)} \widehat \mU = \mS_{\mI}^{(t)} \widehat \mU \widehat{\bf \Lambda},
\end{align}
which is a generalized eigenvalue problem on pair of matrices $(\mS_{\mI, \mA}^{(w)}, \mS_{\mI}^{(t)})$:
columns of $\widehat \mU$ are eigenvectors, and values in the diagonal of $\widehat{\bf \Lambda}$ are eigenvalues \cite{Parlett1998,golub2013matrix}.

As we have a minimization problem, the optimal solution is composed of the $p$ smallest eigenvalues.
They can be computed in $\mathcal{O}((n+k)d^2+d^3)$.

\textbf{Projection dimension.}
We are left with one question: how to choose the projection dimension $p$?
Note that given enough training data ($n \gg d$), which is the case in practice, the total scatter $\mS^{(t)}_\mI$ will be of full rank, that is invertible.
We can rewrite Equation~(\ref{eq:stationarity}) as follows (see Appendix~\ref{app:subspace_dim}):
\begin{align}
    \label{eq:stationarity2}
\underbrace{(\mS_{\mI}^{(t)})^{-1} \mS_{\mI, \mA}^{(b)}}_{=\mS} \widehat \mU &= \widehat \mU \underbrace{(\mI - \widehat{\bf \Lambda})}_{=\widehat{\bf \Lambda}'}\,
\end{align}
which is equivalent to computing eigenvalues $\widehat{\bf \Lambda}'$ of matrix $\mS$.
We set $p$ to the maximum number of non-null eigenvalues we can get, that is:
\begin{align*}
    p&=\rank(\mS)
    = \rank\left((\mS_{\mI}^{(t)})^{-1} \mS_{\mI, \mA}^{(b)}\right)
    \\
    &= \min\left(
        \rank\left(
            (\mS_{\mI}^{(t)})^{-1}
        \right),
        \rank\left(\mS_{\mI, \mA}^{(b)}\right)
        \right)
    \\
    &= \min(d, k-1) = k-1\,.
\end{align*}
As the rank of scatter matrix $\mS_{\mI, \mA}^{(b)}$ is equal to $k-1$ in non degenerated cases.\footnote{It is a sum of $k$ rank-1 matrices that are tied by the mean.}%


\section{Related Work}

\begin{table*}[!ht]
    \footnotesize
    \centering
    \begin{tabular}{lcccccccc}
        \toprule
        \textbf{Model} & \multicolumn{4}{c}{\textbf{1-shot}} & \multicolumn{4}{c}{\textbf{5-shot}} \\
        \cmidrule(lr){2-5} \cmidrule(lr){6-9}
        & \textbf{A} & \textbf{B} & \textbf{C} & \textbf{Avg.}
        & \textbf{A} & \textbf{B} & \textbf{C} & \textbf{Avg.} \\
        Proto $\dagger$                      & $19.3 {\scriptstyle \pm 3.9}$               & $22.7 {\scriptstyle \pm 8.9}$             & $18.9 {\scriptstyle \pm 7.9}$              & $20.3$             & $30.5 {\scriptstyle \pm 3.5}$              & $38.7 {\scriptstyle \pm 5.6}$             & $41.1 {\scriptstyle \pm 3.3}$             & $36.7$ \\
        NNShot $\dagger$                     & $28.5 {\scriptstyle \pm 9.2}$               & $27.3 {\scriptstyle \pm 12.3}$            & $21.4 {\scriptstyle \pm 9.7}$              & $25.7$             & $44.0 {\scriptstyle \pm 2.1}$              & $51.6 {\scriptstyle \pm 5.9}$             & $47.6 {\scriptstyle \pm 2.8}$             & $47.7$ \\
        StructShot $\dagger$                 & $30.5 {\scriptstyle \pm 12.3}$              & $28.8 {\scriptstyle \pm 11.2}$            & $20.8 {\scriptstyle \pm 9.9}$              & $26.7$             & $47.5 {\scriptstyle \pm 4.0}$              & $53.0 {\scriptstyle \pm 7.9}$             & $48.7 {\scriptstyle \pm 2.7}$             & $49.8$ \\
        CONTaiNER $\dagger$                  & $32.2 {\scriptstyle \pm 5.3}$               & $30.9 {\scriptstyle \pm 11.6}$            & $32.9 {\scriptstyle \pm 12.7}$             & $32.0$             & $51.2 {\scriptstyle \pm 5.9}$              & $55.9 {\scriptstyle \pm 6.2}$             & $61.5 {\scriptstyle \pm 2.7}$             & $56.2$ \\
        ~ + Viterbi $\dagger$                & $32.4 {\scriptstyle \pm 5.1}$               & $30.9 {\scriptstyle \pm 11.6}$            & $33.0 {\scriptstyle \pm 12.8}$             & $32.1$             & $51.2 {\scriptstyle \pm 6.0}$              & $56.0 {\scriptstyle \pm 6.2}$             & $61.5 {\scriptstyle \pm 2.7}$             & $56.2$ \\
        \midrule
        \multicolumn{9}{l}{\textbf{Our reproduction on our support sets}} \\
        \midrule
        NNShot                               & $23.9 {\scriptstyle \pm 10.0}$              & $28.2 {\scriptstyle \pm 8.1}$             & $23.0 {\scriptstyle \pm 8.5}$              & $25.0$             & $37.9 {\scriptstyle \pm 6.1}$              & $50.6 {\scriptstyle \pm 6.6}$             & $38.8 {\scriptstyle \pm 3.5}$             & $42.4$ \\
        StructShot                           & $24.6 {\scriptstyle \pm 10.2}$              & $28.2 {\scriptstyle \pm 8.0}$             & $23.4 {\scriptstyle \pm 8.6}$              & $25.4$             & $40.2 {\scriptstyle \pm 6.0}$              & $50.9 {\scriptstyle \pm 6.8}$             & $41.5 {\scriptstyle \pm 4.1}$             & $44.2$ \\
        \midrule
        \multicolumn{9}{l}{\textbf{K-Means with subspace selection using unlabeled dev + train sets}} \\
        \midrule
        \multicolumn{9}{l}{\textbf{Hard clustering (\# \textsc{O}-clusters = 10, \# \textsc{I}-clusters = 1)}}
        \\
        \midrule
        $r_\textsc{O} = NA$                  & $39.5 {\scriptstyle \pm 11.6}$             & $60.3 {\scriptstyle \pm 7.8}$              & $46.6 {\scriptstyle \pm 10.5}$             & $48.8$             & $36.4 {\scriptstyle \pm 10.3}$             & $70.1 {\scriptstyle \pm 4.4}$             & $57.6 {\scriptstyle \pm 6.2}$             & $54.7$ \\
        $r_\textsc{O} = 0.95, 0.96, 0.93$    & $\mathbf{43.5 {\scriptstyle \pm 12.8}}$    & $\mathbf{60.6 {\scriptstyle \pm 6.4}}$     & $45.1 {\scriptstyle \pm 11.3}$             & $\mathbf{49.7}$    & $\mathbf{54.5 {\scriptstyle \pm 13.8}}$    & $69.2 {\scriptstyle \pm 7.8}$             & $60.1 {\scriptstyle \pm 6.3}$             & $\mathbf{61.3}$ \\
        \midrule
        \multicolumn{9}{l}{\textbf{Soft clustering (\# \textsc{O}-clusters = 10, \# \textsc{I}-clusters = 1)}}
        \\
        \midrule
        $r_\textsc{O} = NA$                  & $39.4 {\scriptstyle \pm 11.6}$             & $60.3 {\scriptstyle \pm 7.8}$              & $46.5 {\scriptstyle \pm 10.6}$             & $48.7$             & $35.9 {\scriptstyle \pm 10.5}$             & $70.0 {\scriptstyle \pm 4.4}$             & $57.6 {\scriptstyle \pm 6.2}$             & $54.5$ \\
        $r_\textsc{O} = 0.95, 0.96, 0.93$    & $40.1 {\scriptstyle \pm 11.7}$             & $57.7 {\scriptstyle \pm 13.5}$             & $\mathbf{47.1 {\scriptstyle \pm 11.5}}$    & $48.3$             & $47.1 {\scriptstyle \pm 10.7}$             & $\mathbf{72.3 {\scriptstyle \pm 5.4}}$    & $\mathbf{63.5 {\scriptstyle \pm 5.9}}$    & $61.0$ \\
        \bottomrule
    \end{tabular}
    \caption{
        \footnotesize
        Results for the tag set extension experiments reported in F1-score.
        Results marked with $\dagger$ are taken from \citet{das-2022-container-contrastive}, and are evaluated on different support sets than ours.
        For our evaluation, we generated 10 support sets following the sampling algorithm proposed by \citet{yang-2020-simple-fewshot}.
        Ratio constraints are noted in order of datasets, i.e.\ $r_\textsc{O} = 0.95, 0.96, 0.93$ means that models evaluated on tag set A, tag set B, tag set C use a $0.95$, $0.96$, $0.93$ ratio respectively.
    }
    \label{tab:main-results-tagset-extension}
\end{table*}

\textbf{Few-shot learning.}
A common approach for few-shot learning is to learn a neural network based metric distance \cite[][\emph{inter alia}]{vinyals-2016-matching-networks,snell-2017-prototypical-networks,sung-2018-relation-network}.
Although our approach can also be interpreted as metric learning,
we simplify the process by restricting ourselves to using the euclidean distance after projection in a subspace, where the subspace projection is learned.

In the case of NER, \citet{fritzler-2019-fewshot-prototype} adapted the prototypical network of \citet{snell-2017-prototypical-networks}.
\citet{yang-2020-simple-fewshot} rely on nearest-neighbor classification along with a meta-transition matrix learned from the source task but with simpler IO transitions.
Unfortunately, the later requires to tune a temperature hyper-parameter on the target domain, which is not realistically possible in the few-shot setting where there is no development set.
\citet{das-2022-container-contrastive} extended this approach by fine-tuning on both of the source dataset and target support using a contrastive loss function.
\citet{chen2023promptner} proposed to augment the input with output label information, similarly to the \textsc{GliNER} approach \cite{zaratiana2024gliner}.
Closer to our work, \citet{hou-etal-2020-shot} rely on a target task specific linear projection as proposed by \citet{yoon2019tapnet}, but they cannot benefit from extra unlabeled data.

\textbf{Ratio constraints.}
Using a weak supervision signal in the E step has been known as posterior regularization in the case of generative models \citep{ganchev2010posteriorregularization}.
However, generic application of this framework rely on costly gradient descent to compute the solution of the E step, whereas we propose a polynomial analytical solution for our hard k-means case and an efficient approach based on iterative Bregman projections for the soft case,.
Previous work on ratio constraints for k-means reduced the E step to transportation problems \citep{ng2000constrainedkmeans,bradley2000constrainedkmeans}
but rely on generic algorithms, whereas we propose an efficient algorithm that benefit from the structure of our ratio constraint.

\textbf{Subspace selection.}
Joint k-means and subspace selection is known as \emph{discriminative k-means} \citep{ding-2007-adaptative-dimension-reduction, ye-2007-adaptative-distance-metric-learning-for-clustering, delatorre-2006-discriminative-cluster-analysis, ye-2007-discriminative-k-means}.
We depart from previous work \citep[\emph{e.g.}][]{ding-2007-adaptative-dimension-reduction} by proposing an grounded and well-defined approaches instead of the mere combination of independent steps.

\section{Experiments}

\begin{table*}[!ht]
    \footnotesize
    \centering
    \begin{tabular}{lcccccc}
        \toprule
        \textbf{Model} & \multicolumn{3}{c}{\textbf{1-shot}} & \multicolumn{3}{c}{\textbf{5-shot}} \\
        \cmidrule(lr){2-4} \cmidrule(lr){5-7}
        & \textbf{CoNLL} & \textbf{WNUT17} & \textbf{Avg.}
        & \textbf{CoNLL} & \textbf{WNUT17} & \textbf{Avg.} \\
        Proto $\dagger$                                & $49.9 {\scriptstyle \pm 8.6} $             & $17.4 {\scriptstyle \pm 4.9}$             & $33.7$             & $61.3 {\scriptstyle \pm 9.1}$             & $22.8 {\scriptstyle \pm 4.5}$             & $42.1$ \\
        NNShot $\dagger$                               & $61.2 {\scriptstyle \pm 10.4}$             & $22.7 {\scriptstyle \pm 7.4}$             & $41.9$             & $74.1 {\scriptstyle \pm 2.3}$             & $27.3 {\scriptstyle \pm 5.4}$             & $50.7$ \\
        StructShot $\dagger$                           & $62.4 {\scriptstyle \pm 10.5}$             & $24.2 {\scriptstyle \pm 8.0}$             & $43.3$             & $74.8 {\scriptstyle \pm 2.4}$             & $30.4 {\scriptstyle \pm 6.5}$             & $52.6$ \\
        CONTaiNER $\dagger$                            & $57.8 {\scriptstyle \pm 10.7}$             & $24.2 {\scriptstyle \pm 2.9}$             & $41.0$             & $72.8 {\scriptstyle \pm 2.0}$             & $27.7 {\scriptstyle \pm 2.2}$             & $50.3$ \\
        ~ + Viterbi $\dagger$                          & $61.2 {\scriptstyle \pm 10.7}$             & $27.5 {\scriptstyle \pm 1.9}$             & $44.4$             & $\mathbf{75.8 {\scriptstyle \pm 2.7}}$        & $32.5 {\scriptstyle \pm 3.8}$             & $54.2$ \\
        \midrule
        \multicolumn{7}{l}{\textbf{Hard clustering (\# \textsc{O}-clusters = 10, \# \textsc{I}-clusters = 1)}}
        \\
        \midrule
        $r_\textsc{O} = NA$                            & $65.4 {\scriptstyle \pm 12.0}$             & $20.9 {\scriptstyle \pm  8.2}$             & $43.2$             & $75.7 {\scriptstyle \pm 2.1}$             & $26.5 {\scriptstyle \pm 5.3}$             & $51.1$ \\ 
        $r_\textsc{O} = 0.80, 0.90$                    & $62.6 {\scriptstyle \pm 10.6}$             & $23.6 {\scriptstyle \pm  7.1}$             & $43.1$             & $71.9 {\scriptstyle \pm 3.0}$             & $29.3 {\scriptstyle \pm 3.1}$             & $50.6$ \\ 
        $r_\textsc{O} = 0.85, 0.95$                    & $\mathbf{66.4 {\scriptstyle \pm 13.1}}$    & $\mathbf{28.9 {\scriptstyle \pm  9.4}}$    & $\mathbf{47.7}$    & $\mathbf{75.8 {\scriptstyle \pm 2.6}}$    & $\mathbf{39.0 {\scriptstyle \pm 3.6}}$    & $\mathbf{57.4}$ \\ 
        \midrule
        \multicolumn{7}{l}{\textbf{Soft clustering (\# \textsc{O}-clusters = 10, \# \textsc{I}-clusters = 1)}}
        \\
        \midrule
        $r_\textsc{O} = NA$                            & $65.4 {\scriptstyle \pm 12.0}$             & $20.8 {\scriptstyle \pm  8.1}$             & $43.1$             & $75.7 {\scriptstyle \pm 2.1}$             & $26.5 {\scriptstyle \pm 5.3}$             & $51.1$ \\ 
        $r_\textsc{O} = 0.80, 0.90$                    & $65.6 {\scriptstyle \pm 12.5}$             & $26.6 {\scriptstyle \pm  8.7}$             & $46.1$             & $75.5 {\scriptstyle \pm 2.5}$             & $35.7 {\scriptstyle \pm 3.5}$             & $55.6$ \\ 
        $r_\textsc{O} = 0.85, 0.95$                    & $65.6 {\scriptstyle \pm 12.5}$             & $26.8 {\scriptstyle \pm  8.8}$             & $46.2$             & $75.4 {\scriptstyle \pm 2.5}$             & $35.8 {\scriptstyle \pm 3.5}$             & $55.6$ \\ 
        \bottomrule
    \end{tabular}
    \caption{
        \footnotesize
        Results for the domain adaptation experiments reported in F1-score.
        Results marked with $\dagger$ are taken from \citet{das-2022-container-contrastive} and evaluation of our proposed method is done on the same 10 supports.
        Ratio constraints are noted in the order of datasets, i.e.\ $r_\textsc{O} = 0.80, 0.90$ means that the first dataset, CoNLL, is tested with a ratio of $0.80$ and the second dataset, WNUT17, is tested with a ratio of $0.90$.
    }
    \label{tab:main-results-domain-adaptation}
\end{table*}

We follow previous work \cite{yang-2020-simple-fewshot, das-2022-container-contrastive} and use OntoNotes5 \cite{weischedel-2013-ontonotes} as generic data (news, conversational telephone speech, weblogs, usenet newsgroups, broadcast, talk shows)
together with CoNLL2003 \cite{tjong-kim-sang-de-meulder-2003-conll} and WNUT17 \cite{derczynski-etal-2017-wnut} as specialized data (news and social media, respectively).\footnote{Previous works also evaluate additionally on I2B2 2014 \cite{stubbs-2015-i2b2-2014} for the medical domain.
Despite sharing the support sets they used on this dataset, they do not share the preprocessing steps they employed for sentence segmentation and tokenization, which hinder our comparative evaluation.}

We initialise the model with \texttt{base-bert-cased} \cite{devlin2019bert}.
We pre-train on the source domain for 3 epochs using a $5 \times 10^{-5}$ learning rate with a linear decay.
For consistency, the pre-training is performed with the IO tagging scheme.
If a word is splitted into subtokens, we average the hidden representation of first and last subtokens.

When evaluating in the few-shot settings, we use the train set and development set belonging to each support as unlabeled data.
For k-means, we use 10 iterations of the alternate convex search procedure.
Note that this step takes approximately 6 minutes on CPU when using the full CoNLL dataset.
We have one cluster per entity type (\textsc{I} clusters), and we fix the number of clusters for the \textsc{O} tag to 10.
As we cannot assume to know the true ratio of \textsc{O} tags,
we evaluate our approach with both under- and over-estimations.

\textbf{Prediction.}
Given word $s$ with hidden representation $\vx$, we simply predict the tag associated with the closest cluster:\footnote{\citet{yang-2020-simple-fewshot} rely on Viterbi decoding with a transition matrix learned on pre-training data.
However, it has a temperature parameter which can only be tuned on the test data. Therefore we did not adopt this decoding strategy.}
\begin{align*}
    \hat{y} (\vx) \in \phi\left(\argmin_{j \in[k]} \|\mU^\top \vx- \mU^\top \mC_j\|^2\right).
\end{align*}

\subsection{Few-Shot Settings}

\textbf{Tag set extension.}
This setting evaluates the performance of the model on a set of new tags, but without changing the input data domain.
For a group $T' \subset T \setminus \{\textsc{O}\}$ of tags,
we (1) pre-train the model using only mentions of type $T \setminus T'$ in the training data
and (2) evaluate in a few-shot setting on mention of types $T'$.
We follow \citet{yang-2020-simple-fewshot} and use Ontonotes5 three different sets $T'$, reported in Appendix~\ref{app:tagset-ext-splits}.
We sample 10 support sets for each $T'$ using the algorithm provided by \citet{yang-2020-simple-fewshot}.
We compare to the results of previous works evaluated on their own support sets since those are not publicly available.

\textbf{Domain transfer.}
This setting evaluates the performance of the model on a new set of tags semantically different than those seen during pre-training, and on a different input data source.
To this end, we use Ontonotes5 as a for pre-training, and evaluate few-shot performances on CoNLL2003 and WNUT17.
We use the same support sets as \citet{das-2022-container-contrastive}.

\subsection{Results}

Tables \ref{tab:main-results-tagset-extension} and \ref{tab:main-results-domain-adaptation} summarize the results for the tag set extension and the domain transfer experiments, respectively.
The constrained k-means with subspace selection algorithm performs considerably better than other baseline approaches across all experimental settings, with the biggest improvements observed on the tag set extension setting.

Hard and soft assignments result in very similar performances on the unconstrained version of the algorithm.
It is interesting then to notice that hard assignments version benefits more from the ratio information than the soft assignment one.
In the contrary, the soft assignment version of the algorithm is less sensitive to the ratio constraint and seem to keep a stable performance even with sub-optimal ratio.
We hypothesis that this is due to the fact that without ratio constraints, the hard k-means may stick to incorrect early decisions,
whereas soft-assignations allows to escape them.

Ablation results are given in Appendix~\ref{app:ablation}.

\section{Conclusion}

We propose a novel weakly-supervised algorithm for the few-shot NER.
We evaluate our approach on different scenarios and achieve state-of-the-art results.
Future work could consider applying our approach to learning from partial labels \cite{jin2002partial} and to transductive learning \cite{colombo2023transductive}.

\section*{Limitations}

In practice, it is important to be able to differentiate between succeeding mentions of the same type using \textsc{B} tags.
Unfortunately, including \textsc{B} tags is non-trivial in our approach, and solutions should be considered in future research.
The same limitation happens for inner mentions in the case of nested NER.
Although this is an important limitation of our approach, it is also a limitation of previous work for few-shot NER.

We were unable to compare our approach on the I2B2 dataset as authors did not release unlabeled data using their pre-processing method, nor their pre-processing scripts.

\section*{Acknowledgments}
This work was done while first and last authors were researchers at ISIR in the MLIA team, both funded by the Sorbonne Center for Artificial Intelligence (SCAI).

\bibliography{references}

\appendix


\section{Distance and Projection Weights Equivalence}
\label{app:linear_prototype}
The negative squared Euclidean distance with the prototype can be re-written as:
\begin{align*}
    \evw_j & = - \frac{1}{2} \|\vx - \mC_j\|^2 \\
           & = - \frac{1}{2} (\|\vx\|^2 + \|\mC_j\|^2 - 2 \, \langle \vx, \mC_j \rangle ) \\
           & = \langle \vx, \mC_j \rangle - \frac{1}{2} \|\mC_j\|^2 - \frac{1}{2} \|\vx\|^2
\intertext{where we can define:}
    \vd' &= \begin{bmatrix}
        - \frac{1}{2} \|\mC_1\|^2 \\
        \vdots \\
        - \frac{1}{2} \|\mC_{|T|}\|^2]
    \end{bmatrix}\,,
    \\
\intertext{and:}
    c &= - \frac{1}{2} \|\vx\|^2 \in \R\,,
\intertext{therefore:} 
    \vw    & = \mC \vx + \vd' + c\,.
\end{align*}
and by the constant invariance of the Softmax operation \citep[Proposition 1]{blondel2020fy}, we can show that:
\begin{align*}
    \softmax (\mC \vx + \vd' + c) = \softmax (\mC \vx + \vd')
\end{align*}
where $\mC \vx + \vd'$ is a linear model.

\section{Soft k-Means and Ratio Constraints}
\label{app:appendix-constrained-soft}

In this section, we explain how to compute the E step of soft k-means with the ratio constraint.
The method is based on iterative Bregman projections.
We report the reader to \citep{benamou2015bregmanot} and \citep{censor1997parallel} for an in-depth explanation of this method.

\begin{definition}
    \textbf{(Bregram divergence).}
    Let $f : \R^n \to \R \cup \{\infty\}$ be a strictly convex and continuously differentiable function.
    The $f$-Bregman divergence $D_f : \mathrm{dom} f \times \mathrm{int}(\mathrm{dom} f ) \to \R$ is defined as:
    $$
        D_f (\vp, \vq) = f(\vp) - f(\vq) - \langle \nabla f(\vq), \vp - \vq \rangle.
    $$
\end{definition}

\begin{definition}
    \textbf{(Bregman projection).}
    Let $S$ be a set and $f$ a strictly convex and continuously differentiable function.
    The Bregman projection $\mathrm{Proj}_{f, S}$ is defined as:
    $$
        \mathrm{Proj}_{f, S} (\vq) \in \argmin_{\vp \in S} D_f (\vp, \vq).
    $$
\end{definition}

\begin{definition}
    \textbf{(Iterative Bregman projections).}
    Let $D_f$ be a Bregman divergence and $S = \cap_{i=1}^{k} S^{(i)}$ be a set defined as the intersection of $k$ affine sets $S^{(i)}$.
    We consider problems of the following form:
    $$
        \mathrm{Proj}_{f, S} (\vq) = \argmin_{\vp \in S} D_f (\vp, \vq),
    $$
    where $\vq \in \mathrm{dom} f$ is a given input.
    The iterative Bregman projection algorithm computes a solution of this problem as follows:
    \begin{itemize}
        \item $\vp^{(0)} = \vq$,
        \item $\forall t > 0: \vp^{(t)} = \mathrm{Proj}_{f, S^{(t)}} (\vp^{(t - 1)})$, were we extend the indexing of the sets by $k$-periodicity, i.e.\ $S^{(t + k)} = S^{(t)}$.
    \end{itemize}
    We have $\vp^{(t)} \to \mathrm{Proj}_{f, S} (\vq)$ as $t \to \infty$.
\end{definition}

Now, the constrained soft k-means problem over $\mA$ is defined as:
\begin{align*}
    \argmin_\mA~KL[\mA | \exp(-\mD)]~\text{s.t.}~\mA \in \bigcap_{i=1}^3 S^{(i)}
\end{align*}
This previous problem over $\mA$ can be rewritten as a Bregman projection such that:
\begin{align*}
    \argmin_{\mA \in S} KL[\mA | \exp(- \mD)]
    = \mathrm{Proj}_{-H, S} (\exp(-\mD))
\end{align*}
such that $S = \bigcap_{i=1}^3 S^{(i)}$.
Although the full problem does not have an analytic solution, it can be solved approximately given enough iterations using iterative Bregman projections.
We define two affine sets: $S^{(1)} \cap S^{(2)}$ and $S^{(1)} \cap S^{(3)}$.

The solutions of the projection over each of those sets are:
\begin{align*}
    & \widehat{\mA} \in \mathrm{Proj}_{-H, S^{(1)} \cap S^{(2)}} (\mA) \\
    & \iff \widehat{\emA}_{ij} = 
        \frac{\emZ_{ij} \exp\log \emA_{ij}}{\sum_{j' \in [k]} \emZ_{ij'} \exp\log \emA_{ij'}}\,,
    \\
    &
    \qquad \forall i \in [n], j \in [k],
\end{align*}
and
\begin{align*}
    & \widehat{\mA} \in \mathrm{Proj}_{-H, S^{(1)} \cap S^{(3)}} (\mA) \\
    & \iff \widehat{\emA}_{ij} = \\
    & \begin{cases}
        \frac{\emZ_{ij} n \times r_{\textsc{O}} \exp \log \emA_{ij} }{\sum_{i' \in [n], j' \in \sigma^{-1} (\textsc{O})} \emZ_{i'j'} \exp \log \emA_{i'j'}} & \text{if} \; j \in \sigma^{-1}(\textsc{O}),\\ 
        \emZ_{ij} \exp\log\emA_{ij} & \text{otherwise},
    \end{cases} \\
    & \qquad \forall i \in [n], j \in [k].
\end{align*}
From this, we can derive an iterative algorithm to solve the constrained soft k-means problem over $\mA$ using iterative Bregman projections.

\section{Scatter Matrices}
\label{app:scatter}

Given a cluster assignment matrix $\mA$, we assume $\widehat{\mC}$ are the optimal centroids given by Equation~(\ref{eq:m_step}).
To simplify notation, the dependency on $\mA$ of $\widehat{\mC}$ is not explicitly written.
The total, within-class and between-class scatter matrices are defined as follows:
\begin{align*}
    \mS_{\mI, \mA}^{\text{(t)}}
    &= \sum_{i \in [n]} (\mX_i - \overline{\vx}) (\mX_i - \overline{\vx})^\top
    \\
    \mS_{\mI, \mA}^{\text{(w)}}
    &= \sum_{j \in [k]}\sum_{i\in n[n]} \emA_{ij} (\mX_i - \widehat{\mC}_j) (\mX_i - \widehat{\mC}_j)^\top
    \\
    \mS_{\mI, \mA}^{\text{(b)}}
    &= \sum_{j \in [k]} \sum_{i\in n[n]} \emA_{ij} (\widehat{\mC}_j - \overline{\vx}) (\widehat{\mC}_j - \overline{\vx})^\top
\end{align*}
where $\overline{\vx} = n^{-1}\sum_{i \in [n]} \mX_i$ is the sample mean.
The $\mI$ in the denominator indicates we consider data in the original space.
Note that $\tr(\mS_{\mI, \mA}^{\text{(t)}})$, $\tr(\mS_{\mI, \mA}^{\text{(w)}})$ and $\tr(\mS_{\mI, \mA}^{\text{(b)}})$ corresponds to data, intra-cluster and inter-cluster dispersion.

\subsection{Equality}
\label{app:scatter_eq}

In this section, we prove the following equality:
\begin{align*}
    \mS_{\mI, \mA}^{\text{(t)}}
    &= \mS_{\mI, \mA}^{\text{(w)}} + \mS_{\mI, \mA}^{\text{(b)}}\,. \\
\intertext{
Although this equality in well-known in the hard cluster assignment case \cite[e.g.,][Sec.~14.3.5]{hastie2009ml}, our proof also applies to soft assignments.
An important implication of this equality is that we have:}
    \tr(\mS_{\mI, \mA}^{\text{(t)}})
    &= \tr(\mS_{\mI, \mA}^{\text{(w)}}) + \tr(\mS_{\mI, \mA}^{\text{(b)}})\,,
\end{align*}
meaning that minimizing the intra-cluster dispersion (i.e.\ the k-means objective) is equivalent to maximizing the inter-cluster dispersion (i.e.\ finding well-separated clusters) as the data dispersion is constant.

First, note that as each row of a valid assignment matrix must sum to 1, we can write:
\begin{align*}
    \mS^{\text{(t)}}_{\mI, \mA}
    =& \sum_{i \in [n]} (\mX_i - \overline{\vx}) (\mX_i - \overline{\vx})^\top
    \\
    =&  \sum_{i \in [n]} \underbrace{\left(\sum_{j \in [k]} \emA_{ij}\right)}_{=1} (\mX_i - \overline{\vx}) (\mX_i - \overline{\vx})^\top
    \\
    =&  \sum_{j \in [k]} \sum_{i \in [n]} \emA_{ij} (\mX_i - \overline{\vx}) (\mX_i - \overline{\vx})^\top\,.
    \\
\intertext{We now substract and add $\widehat{\mC_j}$ inside the two terms of the matrix multiplication, and then expand the multiplication:}
    =&  \sum_{j \in [k]} \sum_{i \in [n]} \emA_{ij} \begin{array}[t]{l}(\mX_i - \widehat{\mC}_j + \widehat{\mC}_j - \overline{\vx}) \\ \times (\mX_i - \widehat{\mC}_j + \widehat{\mC}_j - \overline{\vx})^\top\end{array}
    \\
    =&  \underbrace{\sum_{j \in [k]} \sum_{i \in [n]} \emA_{ij} (\mX_i - \widehat{\mC}_j) (\mX_i - \widehat{\mC}_j)^\top}_{=\mS^{\text{(w)}}}
    \\
    &+ \underbrace{\sum_{j \in [k]} \sum_{i \in [n]} \emA_{ij} (\widehat{\mC}_j - \overline{\vx}) (\widehat{\mC}_j - \overline{\vx})^\top}_{=\mS^{\text{(b)}}}
    \\
    &+  \sum_{j \in [k]} \sum_{i \in [n]} \emA_{ij} (\mX_i - \widehat{\mC}_j) (\widehat{\mC}_j - \overline{\vx})^\top
    \\
    &+  \sum_{j \in [k]} \sum_{i \in [n]} \emA_{ij} (\widehat{\mC}_j - \overline{\vx}) (\mX_i - \widehat{\mC}_j)^\top
\end{align*}
We are left with showing that the two last terms are null.

We show that the fourth term is null, the third one follows a similar derivation.
We can move out the term that doesn't depends on $i$ from the sum, and then expand the factorization by $\emA_{ij}$:
\begin{align*}
    &\sum_{j \in [k]} \sum_{i \in [n]} \emA_{ij} (\widehat{\mC}_j - \overline{\vx}) (\mX_i - \widehat{\mC}_j)^\top
    \\
    =& \sum_{j \in [k]} (\widehat{\mC}_j - \overline{\vx}) \sum_{i \in [n]} \emA_{ij} (\mX_i - \widehat{\mC}_j)^\top
    \\
    =& \sum_{j \in [k]} (\widehat{\mC}_j - \overline{\vx}) \left(
        \sum_{i \in [n]} \emA_{ij} \mX_i - \sum_{i \in [n]} \emA_{ij} \widehat{\mC}_j
    \right)^\top
    \\
    =& \sum_{j \in [k]} (\widehat{\mC}_j - \overline{\vx}) \left(
        \sum_{i \in [n]} \emA_{ij} \mX_i - \widehat{\mC}_j\sum_{i \in [n]} \emA_{ij} 
    \right)^\top \,.
    \\
\intertext{If we replace the leftmost occurrence of $\widehat{\mC}_j$ using Equation~\ref{eq:m_step}, we can then see that the second term of the matrix multiplication is equal to the null vector:}
    =& \sum_{j \in [k]} (\widehat{\mC}_j - \overline{\vx}) \\
    &\times\underbrace{\left(
        \sum_{i \in [n]} \emA_{ij} \mX_i - \underbrace{\frac{\sum_{i' \in n} \emA_{i'j} \mX_i}{\sum_{i' \in n} \emA_{i'j}}}_{=\widehat{\mC}_j} \sum_{i \in [n]} \emA_{ij}
    \right)}_{=\vzero}^\top\,,
\end{align*}
which ends the proof.

\subsection{Scatter Matrices and Subspace Projection}
\label{app:scatter_proj}

We now turns to the case were the data is projected into a subspace using matrix $\mU^\top$.
The total scatter matrix in this case can be written as follows:
\begin{align*}
    \mS_\mU^{\text{(t)}}
    &= \sum_{i \in [n]} (\mU^\top \mX_i - \mU^\top \overline{\vx}) (\mU^\top \mX_i - \mU^\top \overline{\vx})^\top
    \\
    &\hspace{-0.5cm}= \sum_{i \in [n]} \mU^\top (\mU^\top \mX_i - \mU^\top \overline{\vx}) (\mX_i - \overline{\vx})^\top \mU
    \\
    &\hspace{-0.5cm}= \mU^\top \left(\sum_{i \in [n]} (\mU^\top \mX_i - \mU^\top \overline{\vx}) (\mX_i - \overline{\vx})^\top \right)\mU
    \\
    &\hspace{-0.5cm}= \mU^\top \mS_{\mI, \mA}^{\text{(t)}}\mU\,.
\end{align*}
Similarly, we have:
\begin{align*}
    \mS_\mU^{\text{(w)}} &= \mU^\top \mS_{\mI, \mA}^{\text{(w)}} \mU \\
    \text{and}\quad
    \mS_\mU^{\text{(b)}} &= \mU^\top \mS_{\mI, \mA}^{\text{(b)}} \mU\,,
\end{align*}
and therefore the following relation trivially holds in the projected case too:
\begin{align*}
    \mS_{\mU, \mA}^{\text{(t)}}
    &= \mS_{\mU, \mA}^{\text{(w)}} + \mS_{\mU, \mA}^{\text{(b)}}\,. \\
\end{align*}

\section{M Step with Joint Subpsace Selection}
\label{app:m_step_subspace}

The solution of the M step is left unchanged when the distance depends on a projection matrix $\mU$.
By first order optimality conditions, $\widehat\mC_j$ is a minimizer if and only if:
\begin{align*}
    \nabla_{\widehat\mC_j}
    \sum_{\substack{i \in [n],\\j \in [k]}} \emA_{ij} \| \mU^\top \mX_i - \mU^\top \widehat\mC_j\|^2
    ]
    &
    = \vzero
    \\
    2 \sum_{i \in [n]}
    \emA_{ij} \mU (\mU^\top \mX_i - \mU^\top \widehat\mC_j)
    &
    = \vzero
    \\
    2 \mU \mU^\top \sum_{i \in [n]}
    \emA_{ij} (\mX_i - \widehat\mC_j)
    &
    = \vzero
    \\
    \sum_{i \in [n]}
    \emA_{ij} (\mX_i - \widehat\mC_j)
    &
    = \vzero
    \\
    \widehat\mC_j &= \frac{\sum_{i \in [n]} \emA_{ij} \mX_i}{\sum_{i \in [n]} \emA_{ij}}.
\end{align*}

\section{Subspace Dimension}
\label{app:subspace_dim}

We derive Equation~(\ref{eq:stationarity2}) as follows.
Remember that stationarity condition, Equation~(\ref{eq:stationarity}), is:
\begin{align*}
    \mS_{\mI, \mA}^{(w)} \widehat \mU &= \mS_{\mI}^{(t)} \widehat \mU \widehat{\bf \Lambda}
    \\
\intertext{By assumption, $\mS_{\mI}^{(t)}$ is of full rank, therefore invertible. We multiplty both sides by its inverse, and rewrite $\mS_{\mI, \mA}^{(w)}$ using Equality~(\ref{eq:scatter_eq}):}
    (\mS_{\mI}^{(t)})^{(-1)}(\mS_{\mI}^{(t)} - \mS_{\mI, \mA}^{(b)}) \widehat \mU &=  \widehat \mU \widehat{\bf \Lambda}
    \\
\intertext{By expanding and re-arranging terms, we obtain:}
    \widehat \mU - (\mS_{\mI}^{(t)})^{(-1)} \mS_{\mI, \mA}^{(b)} \widehat \mU &=  \widehat \mU \widehat{\bf \Lambda}
    \\
    (\mS_{\mI}^{(t)})^{(-1)} \mS_{\mI, \mA}^{(b)} \widehat \mU &= \widehat \mU - \widehat \mU \widehat{\bf \Lambda}
    \\
    \underbrace{(\mS_{\mI}^{(t)})^{(-1)} \mS_{\mI, \mA}^{(b)}}_{=\mS} \widehat \mU &= \widehat \mU \underbrace{(\mI - \widehat{\bf \Lambda})}_{=\widehat{\bf \Lambda}'}\,
\end{align*}
which is equivalent to computing eigenvalues $\widehat{\bf \Lambda}'$ of matrix $\mS$.

\section{Tag Set Extension Splits}
\label{app:tagset-ext-splits}

The list of type $T'$ used for few-shot adaptation are:
\begin{description}
    \item[Group A] \texttt{\{ORG, NORP, ORDINAL, WORK OF ART, QUANTITY, LAW\}}
    \item[Group B] \texttt{\{GPE, CARDINAL, PERCENT, TIME, EVENT, LANGUAGE\}}
    \item[Group C] \texttt{\{PERSON, DATE, MONEY, LOC, FAC, PRODUCT\}}
\end{description}

\section{Ablation Results}
\label{app:ablation}

Results using different unlabeled datasets are given in Table~\ref{tab:main-results-domain-adaptation-unlabeled}.
We use either the full train and dev data as unlabeled data, or only the dev data, or no unlabeled data at all (i.e.\ the E step becomes trivial, and the algorithm reduces to subspace selection).

Results without subspace selection are given in Table~\ref{tab:ablation-no-subspace}.
We observe that subspace selection improves results.

\begin{table*}[!ht]
    \footnotesize
    \centering
    \begin{tabular}{lcccccc}
        \toprule
        \textbf{Model} & \multicolumn{3}{c}{\textbf{1-shot}} & \multicolumn{3}{c}{\textbf{5-shot}} \\
        \cmidrule(lr){2-4} \cmidrule(lr){5-7}
        & \textbf{CoNLL} & \textbf{WNUT17} & \textbf{Avg.}
        & \textbf{CoNLL} & \textbf{WNUT17} & \textbf{Avg.} \\
        Proto $\dagger$                                & $49.9 {\scriptstyle \pm 8.6} $             & $17.4 {\scriptstyle \pm 4.9}$             & $33.7$             & $61.3 {\scriptstyle \pm 9.1}$             & $22.8 {\scriptstyle \pm 4.5}$             & $42.1$ \\
        NNShot $\dagger$                               & $61.2 {\scriptstyle \pm 10.4}$             & $22.7 {\scriptstyle \pm 7.4}$             & $41.9$             & $74.1 {\scriptstyle \pm 2.3}$             & $27.3 {\scriptstyle \pm 5.4}$             & $50.7$ \\
        StructShot $\dagger$                           & $62.4 {\scriptstyle \pm 10.5}$             & $24.2 {\scriptstyle \pm 8.0}$             & $43.3$             & $74.8 {\scriptstyle \pm 2.4}$             & $30.4 {\scriptstyle \pm 6.5}$             & $52.6$ \\
        CONTaiNER $\dagger$                            & $57.8 {\scriptstyle \pm 10.7}$             & $24.2 {\scriptstyle \pm 2.9}$             & $41.0$             & $72.8 {\scriptstyle \pm 2.0}$             & $27.7 {\scriptstyle \pm 2.2}$             & $50.3$ \\
        ~ + Viterbi $\dagger$                          & $61.2 {\scriptstyle \pm 10.7}$             & $27.5 {\scriptstyle \pm 1.9}$             & $44.4$             & $\mathbf{75.8 {\scriptstyle \pm 2.7}}$        & $32.5 {\scriptstyle \pm 3.8}$             & $54.2$ \\

        \midrule

        \multicolumn{7}{l}{\textbf{K-Means with subspace selection using unlabeled dev + train sets}} \\
        \midrule
        \multicolumn{7}{l}{\textbf{Hard clustering (\# \textsc{O}-clusters = 10, \# \textsc{I}-clusters = 1)}}
        \\
        \midrule
        $r_\textsc{O} = NA$                            & $65.4 {\scriptstyle \pm 12.0}$             & $20.9 {\scriptstyle \pm  8.2}$             & $43.2$             & $75.7 {\scriptstyle \pm 2.1}$             & $26.5 {\scriptstyle \pm 5.3}$             & $51.1$ \\ 
        $r_\textsc{O} = 0.80, 0.90$                    & $62.6 {\scriptstyle \pm 10.6}$             & $23.6 {\scriptstyle \pm  7.1}$             & $43.1$             & $71.9 {\scriptstyle \pm 3.0}$             & $29.3 {\scriptstyle \pm 3.1}$             & $50.6$ \\ 
        $r_\textsc{O} = 0.85, 0.95$                    & $\mathbf{66.4 {\scriptstyle \pm 13.1}}$    & $\mathbf{28.9 {\scriptstyle \pm  9.4}}$    & $\mathbf{47.7}$    & $\mathbf{75.8 {\scriptstyle \pm 2.6}}$    & $\mathbf{39.0 {\scriptstyle \pm 3.6}}$    & $\mathbf{57.4}$ \\ 
        \midrule
        \multicolumn{7}{l}{\textbf{Soft clustering (\# \textsc{O}-clusters = 10, \# \textsc{I}-clusters = 1)}}
        \\
        \midrule
        $r_\textsc{O} = NA$                            & $65.4 {\scriptstyle \pm 12.0}$             & $20.8 {\scriptstyle \pm  8.1}$             & $43.1$             & $75.7 {\scriptstyle \pm 2.1}$             & $26.5 {\scriptstyle \pm 5.3}$             & $51.1$ \\ 
        $r_\textsc{O} = 0.80, 0.90$                    & $65.6 {\scriptstyle \pm 12.5}$             & $26.6 {\scriptstyle \pm  8.7}$             & $46.1$             & $75.5 {\scriptstyle \pm 2.5}$             & $35.7 {\scriptstyle \pm 3.5}$             & $55.6$ \\ 
        $r_\textsc{O} = 0.85, 0.95$                    & $65.6 {\scriptstyle \pm 12.5}$             & $26.8 {\scriptstyle \pm  8.8}$             & $46.2$             & $75.4 {\scriptstyle \pm 2.5}$             & $35.8 {\scriptstyle \pm 3.5}$             & $55.6$ \\ 

        \midrule
        \midrule

        \multicolumn{7}{l}{\textbf{Using unlabeled dev set only}} \\
        \midrule
        \multicolumn{7}{l}{\textbf{Hard clustering (\# \textsc{O}-clusters = 10, \# \textsc{I}-clusters = 1)}}
        \\
        \midrule
        $r_\textsc{O} = NA$                            & $65.6 {\scriptstyle \pm 11.1}$             & $22.3 {\scriptstyle \pm  9.6}$             & $43.9$             & $75.1 {\scriptstyle \pm  2.5}$              & $28.6 {\scriptstyle \pm  5.3}$             & $51.8$ \\ 
        $r_\textsc{O} = 0.80, 0.90$                    & $63.1 {\scriptstyle \pm 10.4}$             & $20.9 {\scriptstyle \pm  6.3}$             & $42.0$             & $71.2 {\scriptstyle \pm  3.0}$              & $26.6 {\scriptstyle \pm  2.5}$             & $48.9$ \\ 
        $r_\textsc{O} = 0.85, 0.95$                    & $65.9 {\scriptstyle \pm 12.7}$             & $28.1 {\scriptstyle \pm  8.5}$             & $47.0$             & $75.7 {\scriptstyle \pm  2.3}$              & $38.4 {\scriptstyle \pm  3.5}$             & $57.1$ \\ 
        \midrule
        \multicolumn{7}{l}{\textbf{Soft clustering (\# \textsc{O}-clusters = 10, \# \textsc{I}-clusters = 1)}}
        \\
        \midrule
        $r_\textsc{O} = NA$                            & $65.5 {\scriptstyle \pm 11.1}$             & $22.2 {\scriptstyle \pm  9.5}$             & $43.9$             & $75.1 {\scriptstyle \pm  2.5}$              & $28.5 {\scriptstyle \pm  5.2}$             & $51.8$ \\ 
        $r_\textsc{O} = 0.80, 0.90$                    & $66.3 {\scriptstyle \pm 11.4}$             & $25.4 {\scriptstyle \pm  8.4}$             & $45.8$             & $75.1 {\scriptstyle \pm  2.3}$              & $35.8 {\scriptstyle \pm  4.0}$             & $55.4$ \\ 
        $r_\textsc{O} = 0.85, 0.95$                    & $66.3 {\scriptstyle \pm 11.5}$             & $25.5 {\scriptstyle \pm  8.7}$             & $45.9$             & $75.1 {\scriptstyle \pm  2.3}$              & $36.0 {\scriptstyle \pm  4.1}$             & $55.6$ \\ 

        \midrule
        \midrule

        \multicolumn{7}{l}{\textbf{Using the support sets only (\# \textsc{O}-clusters = 10, \# \textsc{I}-clusters = 1)}} \\
        \midrule
        One iteration                                 & $63.8 {\scriptstyle \pm 8.7}$              & $21.0 {\scriptstyle \pm 10.4}$             & $42.4$             & $73.5 {\scriptstyle \pm 3.7}$               & $32.3 {\scriptstyle \pm 4.3}$              & $52.9$ \\

        \bottomrule
    \end{tabular}
    \caption{
        \footnotesize
        Results for the domain adaptation experiments reported in F1-score.
        Results marked with $\dagger$ are taken from \citet{das-2022-container-contrastive} and evaluation of our proposed method is done on the same 10 supports.
        Ratio constraints are noted in the order of datasets, i.e.\ $r_\textsc{O} = 0.80, 0.90$ means that the first dataset, CoNLL, is tested with a ratio of $0.80$ and the second dataset, WNUT17, is tested with a ratio of $0.90$.
    }
    \label{tab:main-results-domain-adaptation-unlabeled}
\end{table*}

\begin{table*}[!ht]
    \footnotesize
    \centering
    \begin{tabular}{lcccccc}
        \toprule
        \textbf{Model} & \multicolumn{3}{c}{\textbf{1-shot}} & \multicolumn{3}{c}{\textbf{5-shot}} \\
        \cmidrule(lr){2-4} \cmidrule(lr){5-7}
        & \textbf{CoNLL} & \textbf{WNUT17} & \textbf{Avg.}
        & \textbf{CoNLL} & \textbf{WNUT17} & \textbf{Avg.} \\

        Proto $\dagger$                                & $49.9 {\scriptstyle \pm 8.6} $             & $17.4 {\scriptstyle \pm 4.9}$             & $33.7$             & $61.3 {\scriptstyle \pm 9.1}$             & $22.8 {\scriptstyle \pm 4.5}$             & $42.1$ \\
        NNShot $\dagger$                               & $61.2 {\scriptstyle \pm 10.4}$             & $22.7 {\scriptstyle \pm 7.4}$             & $41.9$             & $74.1 {\scriptstyle \pm 2.3}$             & $27.3 {\scriptstyle \pm 5.4}$             & $50.7$ \\
        StructShot $\dagger$                           & $62.4 {\scriptstyle \pm 10.5}$             & $24.2 {\scriptstyle \pm 8.0}$             & $43.3$             & $74.8 {\scriptstyle \pm 2.4}$             & $30.4 {\scriptstyle \pm 6.5}$             & $52.6$ \\
        CONTaiNER $\dagger$                            & $57.8 {\scriptstyle \pm 10.7}$             & $24.2 {\scriptstyle \pm 2.9}$             & $41.0$             & $72.8 {\scriptstyle \pm 2.0}$             & $27.7 {\scriptstyle \pm 2.2}$             & $50.3$ \\
        ~ + Viterbi $\dagger$                          & $61.2 {\scriptstyle \pm 10.7}$             & $27.5 {\scriptstyle \pm 1.9}$             & $44.4$             & $\mathbf{75.8 {\scriptstyle \pm 2.7}}$        & $32.5 {\scriptstyle \pm 3.8}$             & $54.2$ \\
        \midrule
                \multicolumn{7}{l}{\textbf{K-Means with subspace selection}} \\
        \midrule
        \multicolumn{7}{l}{\textbf{Hard clustering (\# \textsc{O}-clusters = 10, \# \textsc{I}-clusters = 1)}}
        \\
        \midrule
        $r_\textsc{O} = NA$                            & $65.4 {\scriptstyle \pm 12.0}$             & $20.9 {\scriptstyle \pm  8.2}$             & $43.2$             & $75.7 {\scriptstyle \pm 2.1}$             & $26.5 {\scriptstyle \pm 5.3}$             & $51.1$ \\ 
        $r_\textsc{O} = 0.80, 0.90$                    & $62.6 {\scriptstyle \pm 10.6}$             & $23.6 {\scriptstyle \pm  7.1}$             & $43.1$             & $71.9 {\scriptstyle \pm 3.0}$             & $29.3 {\scriptstyle \pm 3.1}$             & $50.6$ \\ 
        $r_\textsc{O} = 0.85, 0.95$                    & $\mathbf{66.4 {\scriptstyle \pm 13.1}}$    & $\mathbf{28.9 {\scriptstyle \pm  9.4}}$    & $\mathbf{47.7}$    & $\mathbf{75.8 {\scriptstyle \pm 2.6}}$    & $\mathbf{39.0 {\scriptstyle \pm 3.6}}$    & $\mathbf{57.4}$ \\ 
        \midrule
        \multicolumn{7}{l}{\textbf{K-Means without subspace selection}} \\
        \midrule
        \multicolumn{7}{l}{\textbf{Hard clustering (\# \textsc{O}-clusters = 10, \# \textsc{I}-clusters = 1)}}
        \\
        \midrule
        $r_\textsc{O} = NA$                            & $64.0 {\scriptstyle \pm 14.5}$ & $21.7 {\scriptstyle \pm  7.4}$ & $42.9$ & $72.9 {\scriptstyle \pm 2.9}$ & $22.1 {\scriptstyle \pm 4.9}$ & $47.5$ \\ 
        $r_\textsc{O} = 0.80, 0.90$                    & $60.6 {\scriptstyle \pm 14.4}$ & $25.9 {\scriptstyle \pm  6.4}$ & $43.3$ & $70.5 {\scriptstyle \pm 4.8}$ & $29.3 {\scriptstyle \pm 5.0}$ & $49.9$ \\ 
        $r_\textsc{O} = 0.85, 0.95$                    & $65.2 {\scriptstyle \pm 14.7}$ & $28.2 {\scriptstyle \pm  9.1}$ & $46.7$ & $72.5 {\scriptstyle \pm 3.2}$ & $38.3 {\scriptstyle \pm 2.3}$ & $55.4$ \\ 
        \bottomrule
    \end{tabular}
    \caption{
        \footnotesize
        Ablation results. Constrained and unconstrained k-means, with and without the subspace selection step i.e.\ $\mU = \mI_d$.
    }
    \label{tab:ablation-no-subspace}
\end{table*}

\end{document}